\documentclass[letterpaper]{article} % DO NOT CHANGE THIS
\usepackage{aaai2026}  % DO NOT CHANGE THIS
\usepackage{times}  % DO NOT CHANGE THIS
\usepackage{helvet}  % DO NOT CHANGE THIS
\usepackage{courier}  % DO NOT CHANGE THIS
\usepackage[hyphens]{url}  % DO NOT CHANGE THIS
\usepackage{graphicx} % DO NOT CHANGE THIS
\usepackage{natbib}  % DO NOT CHANGE THIS AND DO NOT ADD ANY OPTIONS TO IT
\usepackage{caption} % DO NOT CHANGE THIS AND DO NOT ADD ANY OPTIONS TO IT
\usepackage{algorithm}
\usepackage{algorithmic}
\usepackage{amsmath}
\usepackage{amssymb}
\usepackage{tabularx}
\usepackage{array}

\usepackage{multirow}
\usepackage{rotating}
\usepackage{booktabs}
\usepackage{pifont}
\newcommand{\xmark}{\ding{55}}%

\usepackage{newfloat}
\usepackage{listings}
\DeclareCaptionStyle{ruled}{labelfont=normalfont,labelsep=colon,strut=off} % DO NOT CHANGE THIS
\floatstyle{ruled}
\newfloat{listing}{tb}{lst}{}
\floatname{listing}{Listing}
\title{TokenPowerBench: Benchmarking the Power Consumption of LLM Inference}

\author{
   Chenxu Niu\textsuperscript{\rm 1}, Wei Zhang\textsuperscript{\rm 2}, Jie Li\textsuperscript{\rm 1}, Yongjian Zhao\textsuperscript{\rm 1}, Tongyang Wang\textsuperscript{\rm 1},\\ Xi Wang\thanks{Corresponding author}\textsuperscript{\rm 3}, Yong Chen\textsuperscript{\rm 1}
}
\affiliations{
\textsuperscript{\rm 1}Texas Tech University\\
\textsuperscript{\rm 2}Texas Advanced Computing Center\\
\textsuperscript{\rm 3}Southeast University\\
chenxu.niu@ttu.edu, wei.zhang@tacc.utexas.edu, \{jie.li, yongjian.zhao, tongyang.wang\}@ttu.edu,\\ xi.wang@seu.edu.cn, yong.chen@ttu.edu
}

\begin{document}

\maketitle

\begin{abstract}

Large language model (LLM) services now answer billions of queries per day, and industry reports show that inference, not training, accounts for more than 90\% of total power consumption. However, existing benchmarks focus on either training/fine-tuning or performance of inference and provide little support for power consumption measurement and analysis of inference. We introduce TokenPowerBench, the first lightweight and extensible benchmark designed for LLM-inference power consumption studies. The benchmark combines (i) a declarative configuration interface covering model choice, prompt set, and inference engine, (ii) a measurement layer that captures GPU-, node-, and system-level power without specialized power meters, and (iii) a phase-aligned metrics pipeline that attributes energy to the prefill and decode stages of every request. These elements make it straightforward to explore the power consumed by an LLM inference run;  furthermore, by varying batch size, context length, parallelism strategy and quantization, users can quickly assess how each setting affects joules per token and other energy-efficiency metrics. We evaluate TokenPowerBench on four of the most widely used model series (Llama, Falcon, Qwen, and Mistral). Our experiments cover from 1 billion parameters up to the frontier-scale Llama3-405B model. Furthermore, we release TokenPowerBench as open source to help users to measure power consumption, forecast operating expenses, and meet sustainability targets when deploying LLM services.

\end{abstract}

\section{Introduction}
\label{sec:intro}
% \section{Introduction}
% % \label{sec:intro}

Large Language Models (LLMs) such as LLaMA-series~\cite{touvron2023llama}, Mixtral~\cite{jiang2024mixtral}, Falcon~\cite{almazrouei2023falcon}, Qwen~\cite{bai2023qwen}, GPT-series~\cite{radford2018improving, radford2019language, brown2020language}, OPT ~\cite{zhang2022opt}, and BERT~\cite{devlin2019bert} have rapidly become foundational infrastructure for modern AI applications. With remarkable capabilities in reasoning, summarization, and text generation~\cite{zhang2019exploring, niu2022kv2vec,xu2025revolution, niu2023psqs, wang2024chatcpu, niu2025energy, wan2024assessment, niu2025iceage, liu2024chatchisel, niu2025rechisel, li2025eda, ye2025chatmodel}, LLMs are now central to large-scale production systems that translate, recommend, and converse across billions of user interactions. This rapid adoption has led universities, national laboratories, and cloud providers to set up dedicated GPU-backed inference services and AI testbeds. For instance, Indiana University~\cite{jetstream2025} now dedicate on-premise GPU clusters exclusively to LLM inference, while many national laboratories~\cite{alcf2025, li2023analyzing} operate AI ``testbeds'' that expose thousands of accelerators to external users. As these services scale, a fundamental systems question remains largely underexplored: ``\textbf{What is the power consumption of serving an LLM prompt?}''

While prior work~\cite{mattson2020mlperf, banbury2021mlperf, reddi2020mlperf} primarily focused on the energy demands of training LLMs, recent evidence shows that the inference process now dominates the energy footprint in large-scale deployments. As LLMs are integrated into countless applications serving billions of users, the cumulative cost of inference is expected to significantly exceed the cost of training. According to a report on the operational lifecycle of LLMs from Amazon Web Services (AWS)~\cite{hutt2019deliver}, inference consumes more than 90\% of the energy consumption. While training a frontier model represents a significant one-time capital expense, inference is a continuous operational expenditure that occurs with every use of the model. This constant demand makes inference the primary driver of computational expense, latency, and energy use. The global LLM market, valued at approximately \$5.6 billion in 2024, is projected to exceed \$35 billion by 2030, with a compound annual growth rate (CAGR) of 36.9\%~\cite{MarketsandMarkets2024}. Meanwhile, the AI inference market is forecast to grow from \$106 billion in 2025 to over \$250 billion by 2030, with a compound annual growth rate of 19.2\%~\cite{MarketsandMarkets2025}. Furthermore, Gartner predicts that by 2028, over 80\% of data center workload accelerators will be dedicated to inference, making a significant shift from the historically training-centric deployments~\cite{Gartner2025}. As shown in the sustainability report of Microsoft~\cite{Microsoft2025}, for a LLM service, electricity expenditure is the single largest component of ongoing operating cost. 
In other words, \textbf{Every Watt Counts as Cost}! Therefore, measuring and optimizing inference power consumption have become essential for researchers and companies to reduce both costs and the energy footprint of AI.

Despite its importance, the community still lacks a comprehensive, reproducible, and scalable benchmarking methodology for measuring and analyzing the power and energy cost of LLM inference across model scales, hardware generations, software stacks, and deployment modes (from single-node to multi-node distributed inference). Existing efforts only partially address this need. For instance, MLPerf Inference benchmark~\cite{reddi2020mlperf} standardizes performance benchmarking across a range of machine learning (ML) tasks but does not systematically capture end-to-end power or normalize energy to LLM-specific service units (e.g., Joules per generated token, per prompt). MLPerf Power benchmark~\cite{tschand2025mlperf} extends in-scope measurements, but its current workflows typically (i) emphasize single-node setups, (ii) focus on modest model sizes relative to frontier state-of-the-art (SOTA) LLMs (e.g., Llama 3 405B-class deployments remain largely uncharacterized), and (iii) often depend on external, high-precision metering equipment that is costly and difficult to replicate across institutional testbeds. Moreover, existing benchmark suites rarely explore the configuration space that practitioners routinely tune in production, such as batch sizing strategies, tensor/pipeline parallelism, context length and quantization levels. All of these configuration settings significantly impact both instantaneous power draw and energy per token in average.

To address this gap, we present \textbf{TokenPowerBench}, the first lightweight and extensible benchmark specifically designed to quantify the power consumption and energy cost of LLM inference. TokenPowerBench features a modular instrumentation layer that integrates vendor telemetry APIs (e.g., GPU/CPU/Memory power sensors), node-level energy sampling, and optional rack- or facility-level measurements when available. A single metrics pipeline lines up every power sample with the two main inference phases (prefill and decode) to allow us to pinpoint exactly where energy is spent. We conduct extensive experiments to systematically analyze the relationship between energy cost and inference-relevant parameters, including batch size, maximum context length, parallelization strategy, and quantization. Our major contributions of this work are:

\begin{itemize}
    \item \textbf{First comprehensive benchmark}: TokenPowerBench is the \textbf{first} open-source framework that couples phase-aware power telemetry with token-level normalization, filling a critical gap left by MLPerf and prior profiling tools.
    \item \textbf{Broad model coverage}: Our initial release profiles \textbf{15+ popular open-source LLMs} (1B–405B parameters), including LLaMA series, Mixtral series, Falcon series, and Qwen series, providing the community with the most comprehensive energy dataset to date.
    \item \textbf{First Parameter-sensitivity Analysis}: Provides fine-grained comparisons of how inference parameters (batch size, context length and quantization) impact energy consumption across decode phases.
    \item \textbf{Ready for frontier SOTA models}: Features a detailed analysis based on a case study of Llama 3.1 405B, a representative example of SOTA LLMs.
\end{itemize}

TokenPowerBench enables systematic comparisons across models, deployment scales, and optimization strategies. It offers actionable insights for researchers, developers, and data center operators seeking to reduce the carbon and operational cost of intelligent services.

\section{Related Work}
\label{sec:related}
% \section{Related Work}
% \label{sec:relatedwork}

\begin{table*}[htbp]
\centering
% \caption{Comparison of LLM Power Benchmarking Approaches}
\label{tab:comparison}
\renewcommand*{\arraystretch}{1.2}
    \setlength{\tabcolsep}{9pt}
\resizebox{1\linewidth}{!}{ 
\begin{tabular}{cccccccccc}
\toprule
\multirow{2}{*}{\textbf{Category}} & \multirow{2}{*}{\textbf{Method}} & \textbf{Node-Level} & \textbf{System-Level} & \textbf{LLM-} & \textbf{Hardware} & \textbf{Cost} & \textbf{Real-World} & \multirow{2}{*}{\textbf{Standardized}} & \textbf{Open} \\
 & & \textbf{Power} & \textbf{Power} & \textbf{Specific} & \textbf{Flexibility} & \textbf{Analysis} & \textbf{Scenarios} & & \textbf{Source} \\
\hline
Benchmark & MLPerf Power@HPCA '25 & \checkmark & \checkmark & \textbf{\xmark} & \textbf{\xmark} & \textbf{\xmark} & \checkmark & \checkmark & \textbf{\xmark} \\
\hline
Benchmark & LLM-Inference-Bench@SC '24 & \checkmark & \textbf{\xmark} & \checkmark & \checkmark & \textbf{\xmark} & \checkmark & \textbf{\xmark} & \checkmark \\
\hline
Measurement & Sustainable NLP@arxiv '25 & \checkmark & \textbf{\xmark} & \checkmark & \textbf{\xmark} & \textbf{\xmark} & \checkmark & \textbf{\xmark} & \checkmark \\
\hline
Measurement & Samsi et al.@HPEC '23 & \checkmark & \textbf{\xmark} & \checkmark & \checkmark & \textbf{\xmark} & \checkmark & \textbf{\xmark} & \checkmark \\
\hline
Estimation & Jegham et al.@arxiv '25 & \textbf{\xmark} & \checkmark & \checkmark & \textbf{\xmark} & \checkmark & \textbf{\xmark} & \textbf{\xmark} & \checkmark \\
\hline
\textbf{Benchmark} & \textbf{TokenPowerBench} & \textbf{\checkmark} & \textbf{\checkmark} & \textbf{\checkmark} & \textbf{\checkmark} & \textbf{\checkmark} & \textbf{\checkmark} & \textbf{\checkmark} & \textbf{\checkmark} \\
\hline
\bottomrule
\end{tabular}
}
\caption{Comparison of LLM Power Benchmarking Approaches
}

\end{table*}

\subsection{Machine Learning and System-Level Power Benchmarks}

MLPerf Power~\cite{tschand2025mlperf} is one of the most comprehensive benchmarks for measuring energy efficiency of machine learning systems. It integrates performance and power measurements using high-precision equipment across edge, datacenter, and cloud systems. However, it treats LLM inference as a generic machine learning inference task and does not account for LLM-specific characteristics such as parallelism techniques and memory optimization. Furthermore, its workflows often assume a static model and dataset configuration, and require expensive power instrumentation setups, limiting accessibility and extensibility for broader LLM deployments.

Green500~\cite{feng2007green500} evaluates the energy efficiency of supercomputers using the HPL benchmark. Although it ranks large-scale systems by FLOPS/Watt, it is not designed for ML or LLM-specific workloads. Other efforts propose power-aware profiling tools for general AI workloads, but they do not capture end-to-end inference flows in token-based generation models.

In contrast, TokenPowerBench focuses specifically on LLM inference and provides phase-aware, token-level measurements aligned with model execution patterns. Our framework enables energy-normalized benchmarking (e.g., Joules per token) and supports reproducible multi-node configurations without requiring external metering hardware.

\subsection{LLM Inference Profiling and Power Analysis}

Recent studies begin to highlight the performance and energy cost of LLM inference. For example, LLM-Inference-bench~\cite{chitty2024llm} is a inference benchmark across diverse hardware but limits its power analysis to the accelerators themselves. Poddar et al.~\cite{poddar2025towards} analyze the energy consumption of large transformer models during inference, often using cloud GPUs (e.g., A100, H100) and focusing on distillation or quantization strategies. However, these studies are typically limited to fixed setups, lack systematic benchmarking frameworks, and do not explore parameter space (e.g., batch size, context length) or cluster-wide energy variation. Samsi~\cite{samsi2023words} propose a more focused academic approach using direct hardware telemetry (nvidia-smi/DCGM~\cite{pynvml}) to measure only the accelerator's power consumption. Some works~\cite{jegham2025hungry} investigate the carbon footprint of foundation models, yet mostly focus on training-phase emissions rather than inference.

TokenPowerBench addresses this gap by offering a lightweight, extensible framework to benchmark and analyze energy efficiency across LLM sizes, hardware generations, and deployment modes. It introduces a normalized metric suite (Joules/token, power imbalance, energy-delay product) and automates parameter sweeps over key inference configuration dimensions.

\subsection{System-Level Power Instrumentation}

A number of tools and frameworks exist for system-level power monitoring~\cite{li2020monster, stefanov2021review}. NVIDIA’s NVIDIA Management Library (NVML), Data Center GPU Manager (DCGM) and ``nvidia-smi'' provide GPU-level power telemetry, while Intel’s Running Average Power Limit (RAPL)~\cite{Intel2023RAPL} interface supports package-level CPU/DRAM energy readings. Cluster-wide monitoring systems like Intelligent Platform Management Interface (IPMI)~\cite{ipmi_spec}, Redfish~\cite{DMTF2023Redfish}, and rack-level PDUs offer coarse-grained wall power data. 

% Projects like Accelergy~\cite{wu2019accelergy} aim to model energy use via simulation or software emulation.

TokenPowerBench integrates vendor-native telemetry where available and aligns measurements with LLM inference phases. Our framework supports real-time sampling at multiple granularity levels (GPU, node, rack) and performs time-correlated breakdown across prefill, decode, and idle phases to enable fine-grained attribution of power consumption.

\subsection{Positioning Against Standardized Benchmarks}

Table~\ref{tab:comparison} contrasts TokenPowerBench with the main benchmark and measurement efforts proposed to date for LLM power analysis. TokenPowerBench is the first framework that combines LLM-specific coverage (dense and MoE models up to LLaMA-3-405B), component-level and power measurement without specialized meters, and a fully open-source implementation—thereby filling every gap highlighted in the comparison.

TokenPowerBenchis not a replacement for MLPerf Power or any other benchmark but rather a complementary, agile tool. While MLPerf answers the question ``Which hardware is more efficient on a standard task?'', TokenPowerBench answers the question ``\textbf{What is the real-world energy cost of running my massive, distributed model with my specific configuration on my available hardware?}''. Its lightweight nature makes it the ideal solution for this practical, operator-centric measurement problem.

\section{TokenPowerBench Architecture}
\label{sec:architecture}
% \section{TokenPowerBench Architecture}
% \label{sec:method}

\begin{figure*}[ht]
  \centering
  \includegraphics[width=\linewidth]{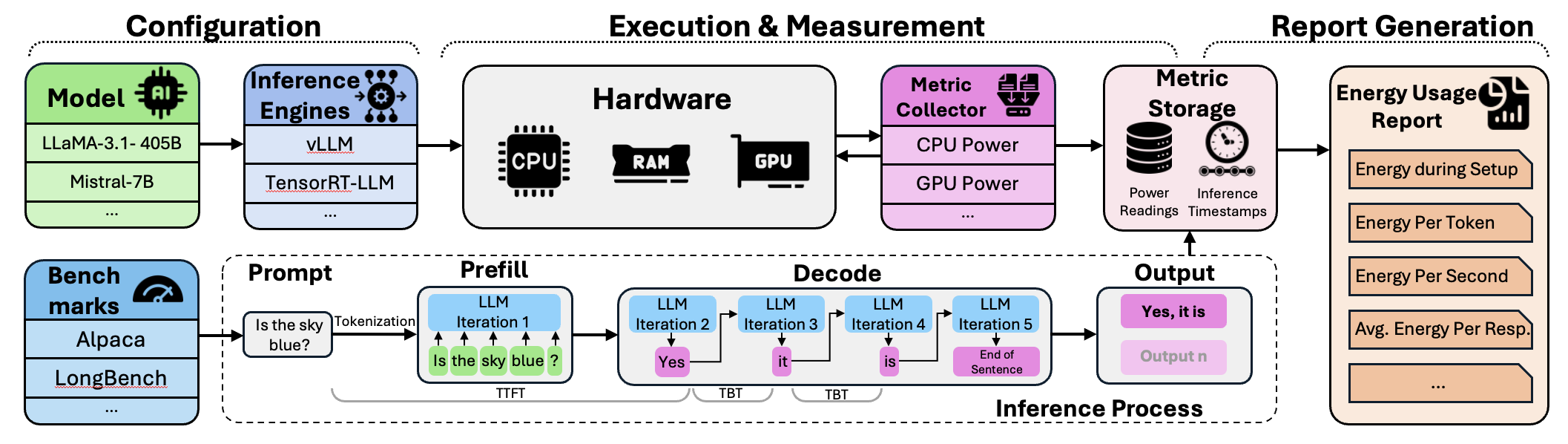}
  \caption{Overview of the TokenPowerBench Architecture}
  \label{fig:architecture}
\end{figure*}

Benchmarking LLM inference power consumption is fundamentally different from benchmarking throughput or latency. Unlike traditional performance metrics, energy consumption is shaped by hardware heterogeneity, software stack complexity, and workload dynamics. All of the above settings vary significantly across deployments. Existing LLM benchmarks offer limited support for this variability, often assuming fixed configurations, static model sizes, and expensive instrumentation.

As shown in the Figure \ref{fig:architecture}, \textbf{TokenPowerBench} overcomes these limitations through a three-layer architecture: Configuration, Execution \& Measurement, and Report Generation. It offers a modular, configurable, fully reproducible, and scalable benchmarking methodology tailored to modern LLM inference. 

\subsection{Configuration}

% The choice of LLMs is critical for the relevance and comparability of any benchmark. We selected a suite of models to provide comprehensive coverage across three key dimensions: architectural diversity, scale, and accessibility. Our goal is to move beyond single-model studies to characterize power consumption trends across the families of models currently deployed.

The first task of TokenPowerBench is to help users set up the test environment by selecting the model to run, the inference engine to use, and prompts to feed into it. To keep this step lightweight, TokenPowerBench exposes three plug-and-play modules: a \textbf{model pool} (choose any supported LLM), a \textbf{prompt-dataset menu} (select from Alpaca, LongBench, or custom prompts), and an \textbf{inference-engine selector} (vLLM, TensorRT-LLM, Transformers, or DeepSpeed).

\begin{itemize}
    \item \textbf{Model Pool}: We include models with different underlying architectures to capture how design choices impact energy profiles. This includes standard decoder-only transformers (e.g., the LLaMA series) and Mixture-of-Experts (MoE) models (e.g., Mixtral), which exhibit distinct computational patterns, particularly in parameter activation and memory access. The benchmark covers a broad spectrum of model sizes, from smaller models with fewer than 1 billion parameters--suitable for a single consumer-grade GPU--to frontier-scale models like Llama 3-405B, which require multi-GPU and/or multi-node distributed inference. This allows us to study how energy consumption sacles with increasing model size and hardware complexity.
    
    % \item \textbf{Accessibility:} All selected models are open-source and widely available through platforms like Hugging Face. This ensures that our experiments are fully reproducible by the broader community, a core tenet of TokenPowerBench.
    \item \textbf{Prompt-dataset memu}: Prompt length and style can significantly affect energy consumption during inference. To capture this variability, TokenPowerBench provides built-in support for two representative datasets: Alpaca, featuring short, chat-style prompts, and LongBench, which includes extended contexts of up to 10k tokens. Users can also supply custom data in CSV or JSON format, including arbitrary lists of prompts, code snippets, or full-text articles.
    \item \textbf{Inference Engines}: TokenPowerBench supports four widely used engines on a single node: vLLM~\cite{kwon2023vllm}, TensorRT-LLM~\cite{vaidya2023tensorrtllm}, DeepSpeed~\cite{aminabadi2022deepspeed}, and Transformers~\cite{face2024transformers}. To accommodate frontier-scale models such as Llama3-405B--which requires at least 780 GB of FP16 memory--we integrate support for distributed inference using the Ray framework. TokenPowerBench automatically launches Ray services across nodes and partitions model weights accordingly. Once initialized, inference engines are executed on Ray, and TokenPowerBench continues to collect power consumption data using a consistent, unified methodology.

\end{itemize}

Beyond the choice of model, inference engine, and dataset, the specific configuration of the inference service is a primary determinant of power consumption. TokenPowerBench is designed to explore the multi-dimensional configuration space that practitioners need to navigate. Instead of reporting a single performance number, it systematically varies key parameters to build a detailed power profile across different operational conditions. The primary axes of our benchmark scenarios are:

\begin{itemize}
    \item \textbf{Hardware Configuration:} We define scenarios for single-GPU workstations, single-node multi-GPU servers (e.g., 8x H100s), and multi-node clusters, enabling direct comparisons of power draw in scale-up versus scale-out deployments.
    \item \textbf{Parallelism Strategy:} For large models, we evaluate the energy implications of different distribution strategies--primarily \textbf{Tensor Parallelism (TP)} and \textbf{Pipeline Parallelism (PP)}. TP partitions model layers across GPUs, increasing inter-GPU communication, while PP stages layers sequentially, introducing potential pipeline bubbles. These trade-offs manifest in distinct patterns of GPU utilization, idle time, and network power draw.
    \item \textbf{Workload Parameters:} We investigate the impact of dynamic request-level variables:
        \begin{itemize}
            \item \textbf{Batch Size:} We sweep from batch size 1 (latency-critical interactive use) to large batches (throughput-oriented offline processing) to evaluate power consumption and energy costs.
            \item \textbf{Context Length:} We vary the number of input prompt tokens to analyze the energy cost of the computationally-intensive prefill stage versus the memory-bandwidth-bound decode stage. 
            \item \textbf{Quantization:} We benchmark various numerical formats (e.g., FP16, FP8). While quantization reduces memory footprint and can accelerate computation, its overall impact on system-wide energy consumption remains a key question.
        \end{itemize}
\end{itemize}

\begin{figure*}[ht]
  \centering
  \includegraphics[width=\linewidth]{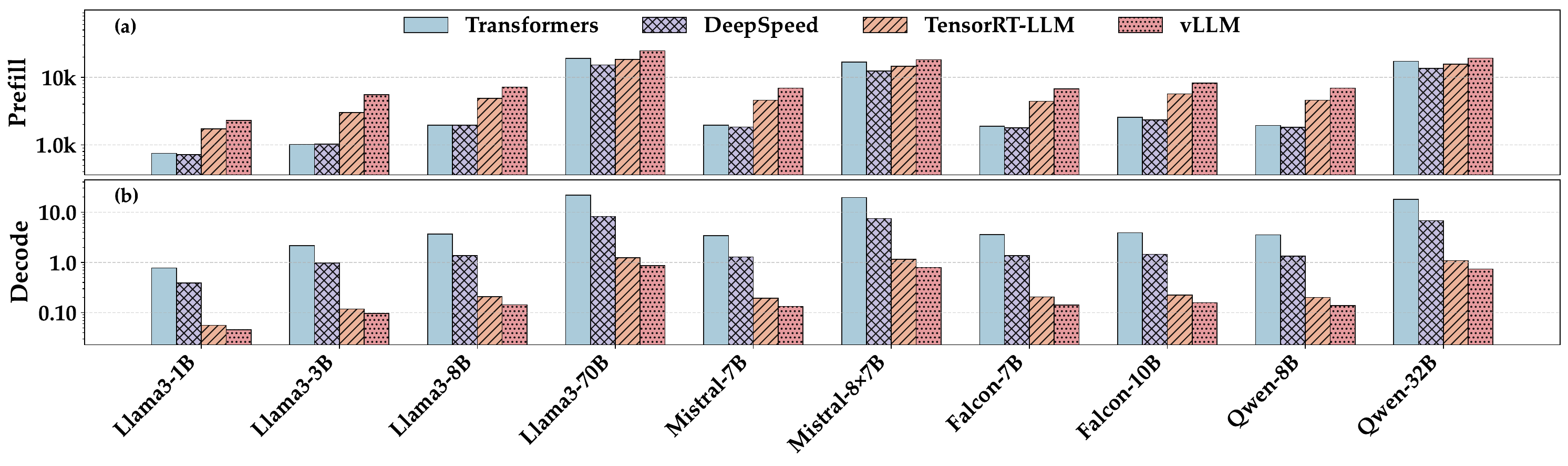}
  \caption{Prefill Energy (a) and Decode Energy per Token (b) Across Models and Inference Engines}
  \label{fig:gpu_per_token}
\end{figure*}

By sweeping through these scenarios, TokenPowerBench provides a holistic view of how operational choices, from hardware provisioning to software optimization, affect the energy efficiency of LLM inference.

\subsection{Execution and Measurement}

\paragraph{1) Spatial view: where the power is spent.}
During each run, TokenPowerBench samples power consumption from all major node components: \emph{GPU}, \emph{CPU}, \emph{DRAM}, and--when sensors are available--the network interface and fan tray. GPU power is collected via NVML/DCGM; CPU and DRAM power via Intel RAPL; and full-node power via IPMI or a rack-mounted PDU. Storing these readings side-by-side enables insights such as GPUs typically account for over 60\% of total energy use, while fans contribute only a few percent. Because all telemetry streams share the same timestamp, we can also sum them precisely to compute the total wall energy for each request.

\paragraph{2) Temporal view: when the power is spent.}
The same logger records two key stages of every inference: \textit{prefill}, when the model reads the input tokens, and
\textit{decode}, when it produces new tokens. Each power sample is tagged with the stage that is active at that moment. After the run we integrate these tagged samples to obtain two clear numbers: energy consumed during prefill and energy consumed during decode. This separation reveals, for instance, that long prompts raise the prefill share, while large batch sizes increase the decode share. By combining spatial and temporal perspectives, we can determine both \emph{which component} and \emph{which stage} accounts for each joule consumed during LLM inference.

% \begin{equation}
% E_{\text{PowerInput}} = E_{\text{GPU}}+E_{\text{CPU}}+E_{\text{DRAM}}+E_{\text{Others}}
% \end{equation} 

\begin{align}
E_{\text{total}} &= E_{\text{Prefill}} + E_{\text{Decode}} \\
                 &= E_{\text{GPU}}+E_{\text{CPU}}+E_{\text{DRAM}}+E_{\text{Others}}
\end{align}

\subsection{Report Generation}

When an experiment ends, TokenPowerBench collects the log data and turns the raw time-stamped samples into a compact summary. First, it integrates the GPU, CPU, DRAM, and wall-plug traces over the two stages of inference including prefill and decode stages, so the user can see at a glance how many joules each component consumed in each part of the process. Users can get the data such as energy per token, energy per response, energy per second, peak powe, and energy consumed during prefill stage. Because every value is computed directly from the aligned samples, there is no need for post-hoc scaling or hand calculations.

The tool then writes the results in three complementary formats.
A CSV file holds the numbers most people plot with Pandas or Excel; a matching JSON file fits easily into automated dashboards;  If the user supplies electricity price and a regional carbon factor, the same pass also converts kilowatt-hours into dollar cost and $CO_2$ equivalents, so budget and sustainability discussions start from the same sheet as the technical metrics. Because every run follows the same pipeline, reports produced on different days or clusters line up without extra scripting, making it straightforward to track regressions or show savings after an optimization.

\begin{figure*}[ht]
  \centering
  \includegraphics[width=\linewidth]{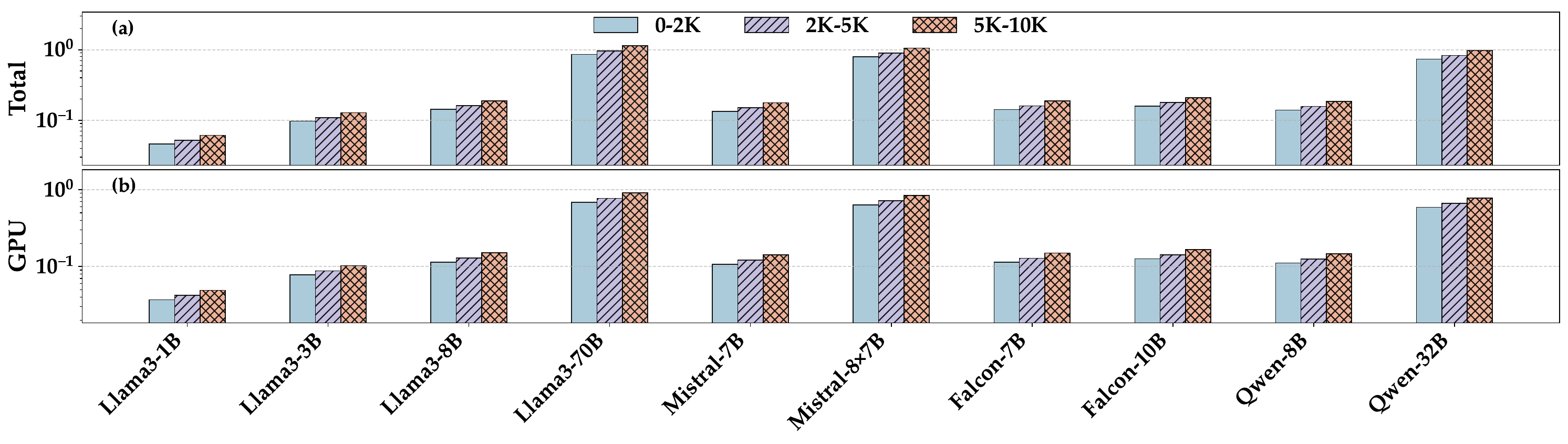}
  \caption{ Comparison of Total (a) and GPU (b) Energy per Token Across Models at Varying Context Lengths}
  \label{fig:context}
\end{figure*}

\section{Evaluation}
\label{sec:eva}
% \section{Evaluation}
% \label{sec:eva}

To demonstrate the utility and effectiveness of TokenPowerBench, we conduct an in-depth case study on a representative Nvidia H100 GPU cluster. Our evaluation is designed to showcase how the benchmark can be used to derive actionable insights into the performance, energy consumption, and cost of deploying large-scale LLMs.

\subsection{Experimental Setup}

\subsubsection{Hardware}

The experiments were performed on a 8-node GPU cluster, each node is equipment with 4 NVIDIA H100 GPUs (94 GB memory each), paired with two Intel Xeon Gold 6426Y CPUs (16 cores, 32 threads each) and 512 GB of RAM.

\subsubsection{Prompt Datasets}

We evaluate our benchmark on two datasets: Alpaca~\cite{alpaca} and LongBench~\cite{bai2023longbench}. Alpaca contains 52,002 prompts generated by OpenAI's text-davinci-003 engine~\cite{openai2022textdavinci003}. LongBench is an open-source benchmark and has longer prompts in average.

% consisting of the following datasets: HotpotQA~\cite{yang2018hotpotqa}, WikiMultiHopQA~\cite{ho2020constructing}, MuSiQue~\cite{trivedi2022musique}, DuReader~\cite{he2017dureader}, NarrativeQA~\cite{kočiský2017narrativeqareadingcomprehensionchallenge}, qasper~\cite{dasigi2021datasetinformationseekingquestionsanswers}, QMSum~\cite{zhong2021qmsumnewbenchmarkquerybased}, VCSUM~\cite{wu2023vcsumversatilechinesemeeting}, TriviaQA~\cite{joshi2017triviaqalargescaledistantly}, SAMSum~\cite{gliwa2019samsum}, Multi-News~\cite{fabbri2019multinewslargescalemultidocumentsummarization} and RepoBench~\cite{liu2023repobenchbenchmarkingrepositorylevelcode}. It has longer prompts in average.

\subsection{Cross-Model Power Consumption Benchmarking of LLMs}

We tested all popular LLM models, including Llama3 - 1B, 3B, 8B, 70B and 405B; Mistral 7B, 24B, 8$\times$7B, 8$\times$22B; Qwen 8B, 32B and 480B; Falcon 7B, 10B and 180B. The representative results as shown in the following table. Please check the total results in the supplementary material.

% The most important metrics to measure energy efficiency is ``energy per token''. 
Figure~\ref{fig:gpu_per_token} compares energy consumed by prefill stage and total energy per token for 10 open-source models (dense and MoE) served by four mainstream engines: Transformers, DeepSpeed-Inference, TensorRT-LLM, and vLLM on a single node with 4 H100 GPUs. Within each LLM family, energy rises faster than the parameter count. For LLaMA-3, moving from 1 B to 70 B parameters increases energy per token by $7.3\times$, even though parameter count grows $70\times$. This super-linear trend confirms that larger models pay an extra cache-bandwidth and memory-traffic penalty beyond pure FLOPs.

\paragraph{Dense versus MoE.}
Mixtral-8$\times$7B consumes roughly the same energy per token as a dense 8B model while delivering quality closer to a 56 B dense model. The sparse routing that activates only two experts per token cuts token energy by \(2\text{–}3\times\) compared with dense models of similar emergent accuracy.

\paragraph{Engine impact.}
Across \emph{all} models, TensorRT-LLM and vLLM consume 3 times more than DeepSpeed and Transformers in the prefill stage. However, TensorRT-LLM and vLLM reduce energy per token by \(25\text{–}40\%\) relative to Transformers engine due to the optimization techniques they adopt. DeepSpeed-Inference sits between the two extremes, showing that software optimization alone can rival architecture-level gains.

\subsection{Parameter-sensitivity Analysis}

\begin{figure}[ht]
  \centering
  \includegraphics[width=\linewidth]{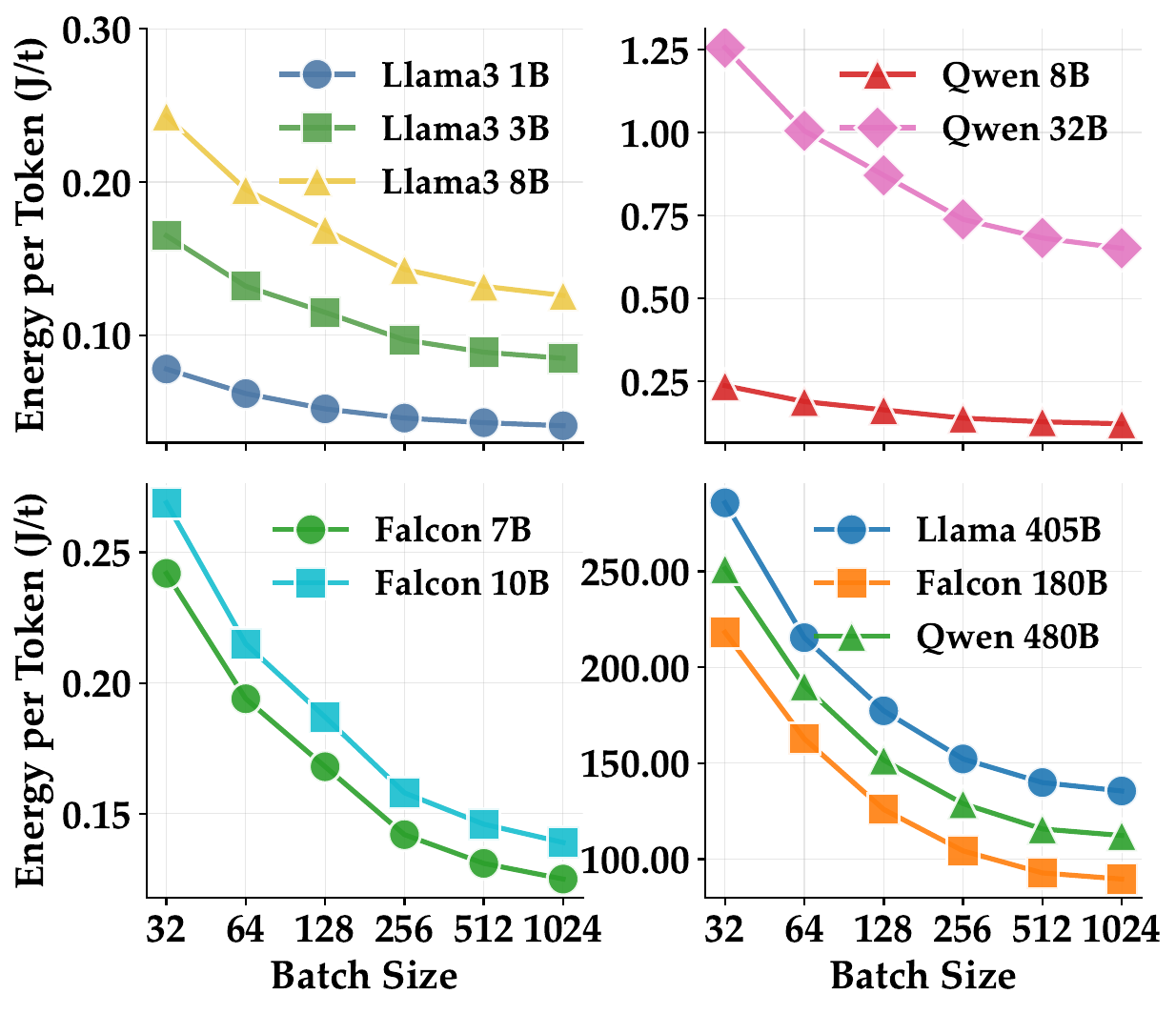}
  \caption{Comparison of Total Energy per Token Across Models at Varying Batch Sizes}
  \label{fig:batch}
\end{figure}

\subsubsection{Batch Size}

Figure~\ref{fig:batch} plots energy per token for batch sizes from 32 to 1024 on six model families, including Llama, Falcon, Mistral Qwen families. For every family, enlarging the batch lowers energy per token at first, because the fixed set-up and kernel-launch overheads are spread over more tokens. The steepest drop appears between 32 and 256, where most GPUs move from \(<50\%\) to nearly full utilization; the 70 B model, for example, cuts per-token energy by about 25\% in this range.  

Beyond batch 256 the curve flattens: power draw stays roughly constant while additional tokens add proportionally less work, so further efficiency gains are modest. Here batch sizes up to 1024 still fit in memory, so energy per token continues to fall, though at a slower rate, giving an overall two-to-three-fold spread between the smallest and largest batches.

% \begin{figure*}[ht]
%   \centering
%   \includegraphics[width=\linewidth]{images/vllm_context_length_energy_comparison.pdf}
%   \caption{ Comparison of Total (a) and GPU (b) Energy per Token Across Models at Varying Context Lengths}
%   \label{fig:context}
% \end{figure*}

\subsubsection{Context Length}

Figure~\ref{fig:context} reports energy (total and gpu) per token for ten models at three prompt length range : 0-2K, 2K to 5K, and 5K to 10K tokens. Across all models, energy grows steadily as the prompt gets longer because the prefill stage must process every input token while the compute cost of each new output token stays the same.  

For the largest dense model we tested (Llama3 70B), the jump from 2K to 10 K tokens raises energy per token by roughly a factor of three; medium models (e.g., Llama 3 8B, Mistral 7B) see a smaller but still clear rise.  The pattern is nearly identical in the GPU-only trace and in the node-level trace, confirming that most of the extra power is drawn by the accelerators rather than by the host CPU or memory.

Longer prompts therefore hurt energy efficiency in two ways: they increase the joules spent before the first output token appears, and they lower overall throughput because GPUs stay busy on attention over a larger context window.

\subsubsection{Parallelism Strategy Study (TP / PP) — SOTA LLMs}

\begin{figure}[ht]
  \centering
  \includegraphics[width=\linewidth]{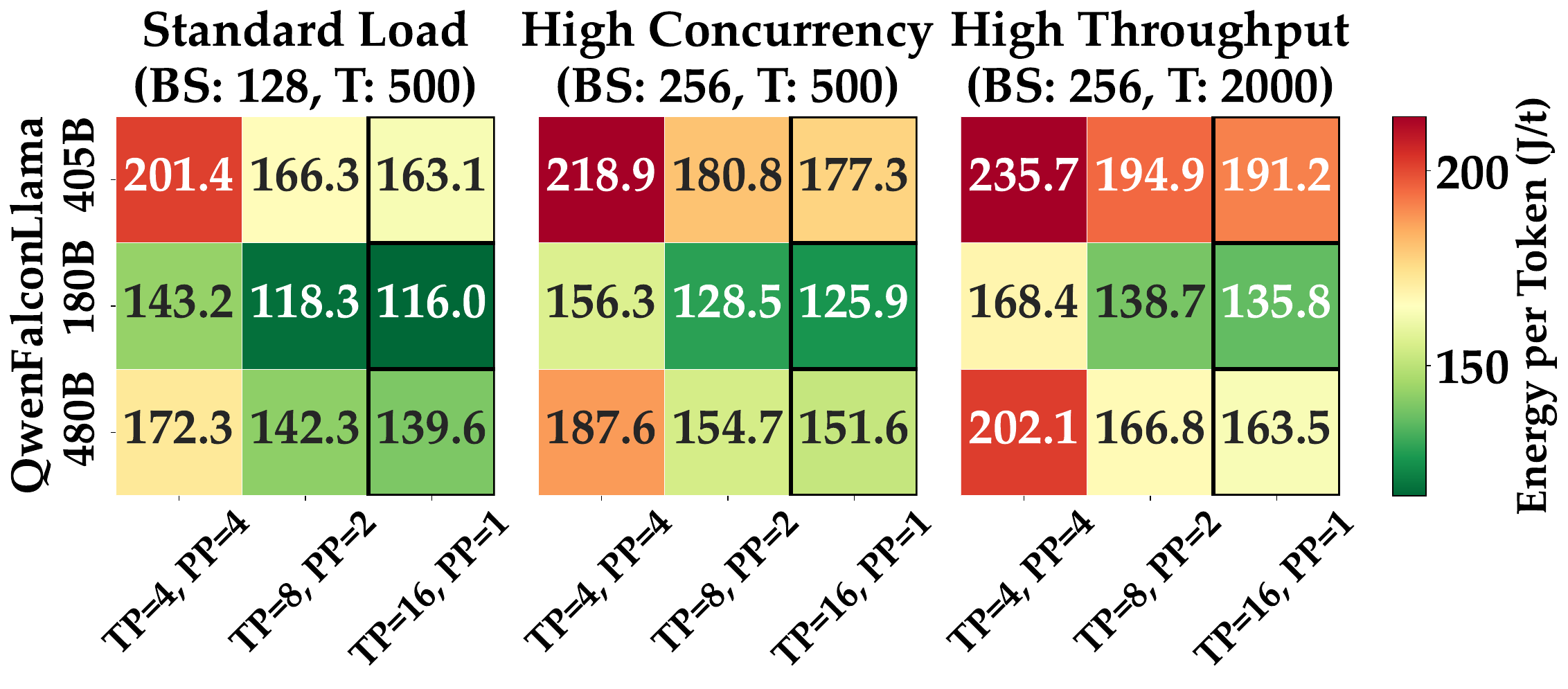}
  \caption{Heatmap of Energy per Token of SOTA models}
  \label{fig:heatmap}
\end{figure}

Three workload configurations were used during the token generation stage to simulate the workloads of real-world scenarios: Standard Load, High Concurrency, and High Throughput.

The heatmap figure presents a heatmap for three SOTA models: Llama 3 405B, Falcon180B, and Qwen 480B on 16 H100 GPUs under three workload profiles. For each model we tested three ways to split work across the GPUs: a balanced mix of tensor and pipeline parallelism with TP 4 and PP 4, a tensor-heavy mix with TP 8 and PP 2, and pure tensor parallelism with TP 16 and PP 1. Green cells in the figure mark lower energy use.

Across every workload the \textbf{pure tensor parallelism} setting  delivers the best energy efficiency because long pipelines leave some GPUs idle. The gap between the best and worst split widens as the workload grows from about 40 J/token under Standard Load to more than 60 J/token under High Throughput. The results show that parallelism tuning matters most in heavy, batch-oriented jobs.

\subsubsection{Quantization - A Case Study of Llama 3 405B (FP16 vs FP8)}

\begin{figure}[ht]
  \centering
  \includegraphics[width=\linewidth]{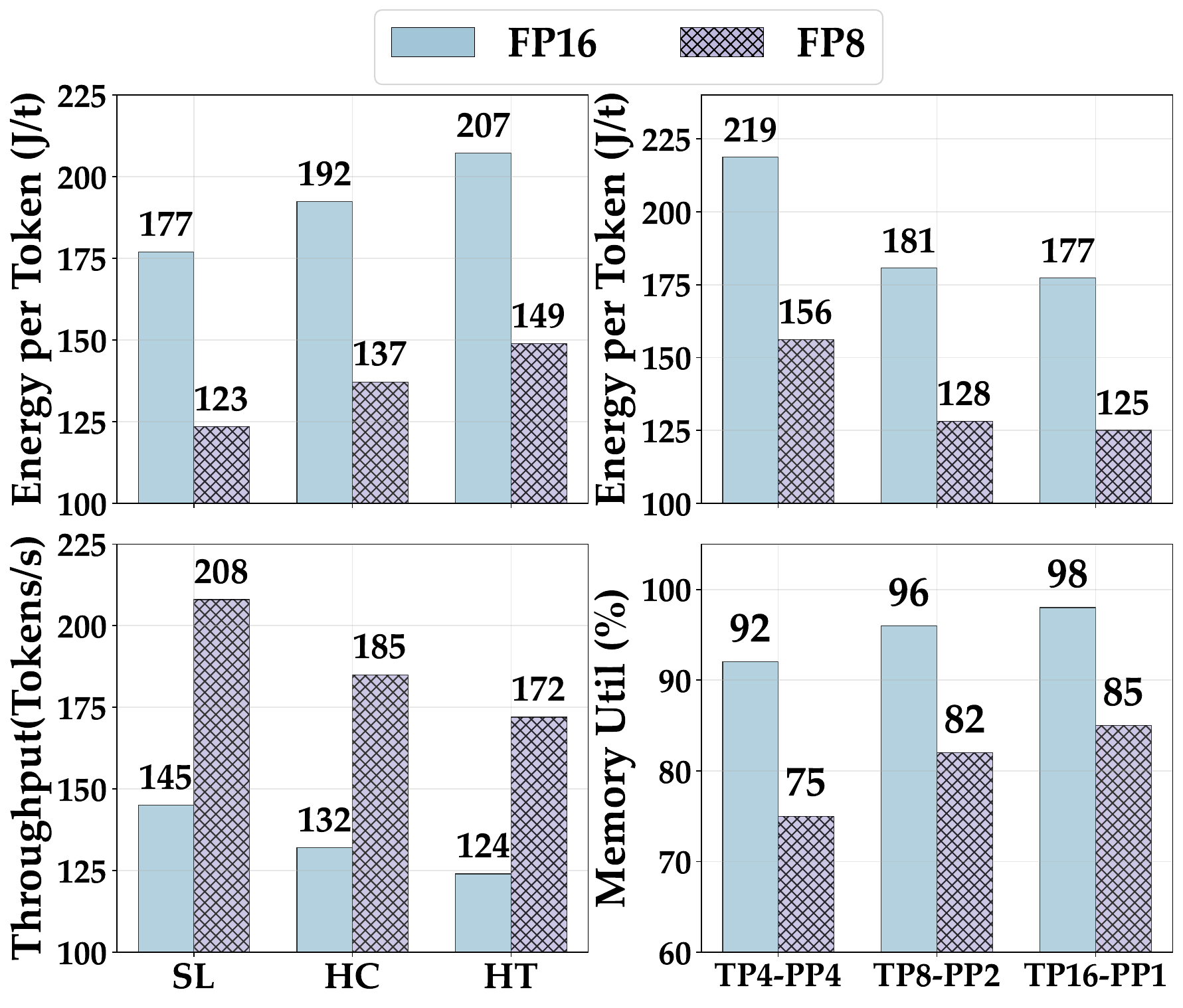}
  \caption{Comparison of FP16 and FP8 Performance for the LLaMA 3 405B Model Across Varying Workloads and Parallelism Levels. (SL: Standard Load, HC: High Concurrency, HT: High Throughput)}
  \label{fig:quantization}
\end{figure}

Figure \ref{fig:quantization} compares full-precision FP16 (16-bit Floating Point) inference with FP8 (8-bit Floating Point) weight quantization for Llama 3 405B. Across the three workload profiles: Standard Load, High Concurrency, and High Throughput, quantization cuts energy per token by roughly 30\%.  The saving is consistent regardless of the traffic pattern because the reduced-precision weights lower both memory traffic and arithmetic cost in every phase of generation. Measured per batch, total energy falls from about 45 kJ to 32 kJ under the heaviest load, mirroring the token-level reduction.

Performance also benefits by the quantization. FP16 increases effective memory-bandwidth utilization by 13–17 percentage, but FP8 raises end-to-end throughput from about 48-63 tokens/s in the largest batch setting, without noticeable accuracy loss in our prompt set. Together, these results confirm that low-precision formats can deliver a double dividend lower power and higherspeed, when the underlying hardware and kernels fully support them.

\section{Conclusion and Future Work}
\label{sec:con}
% \section{Conclusion and Future Work}
% \label{sec:con}

TokenPowerBench’s core contribution is a lightweight, extensible, and reproducible framework for benchmarking the power and energy consumption of LLM inference across diverse hardware configurations, model families, and workload patterns. We briefly summarize the core aspects of our benchmark design here.

\textbf{Energy-normalized inference metrics.} To enable meaningful comparison of inference systems, TokenPowerBench introduces a set of energy-centric metrics including Joules per token, Joules per response and instantaneous power draw. These metrics are aligned with the internal execution phases of transformer-based LLMs (prefill and decode), enabling fine-grained attribution of energy costs. Unlike traditional performance metrics like latency or throughput, these energy metrics directly reflect sustainability and cost-efficiency, which are critical concerns in modern AI infrastructure.

\textbf{Reproducible measurement methodology.} Power measurement in LLM inference is highly sensitive to hardware access level and instrumentation fidelity. TokenPowerBench defines a three-level measurement model—from GPU-only telemetry to full-system power monitoring using IPMI and rack-level PDUs—allowing users to adopt the benchmark in environments ranging from user-space workstations to institutional testbeds. We pair this with a declarative configuration harness that ensures repeatability across experiments.

\textbf{Scalable scenario design.} LLM inference workloads vary widely based on system scale, model size, and deployment constraints. TokenPowerBench systematically explores key configuration dimensions, including batch size, context length, quantization format, parallelism strategy to expose how each factor impacts energy usage. We support scaling from single-GPU inference to multi-node distributed serving, capturing both component-level breakdowns and cluster-wide energy imbalances.

% \textbf{Insights for sustainable AI.} As LLMs are increasingly deployed in production environments serving millions of users, understanding the power implications of design choices—model selection, hardware provisioning, workload scheduling—is no longer optional. TokenPowerBench bridges the gap between performance-focused evaluation and energy-aware system design, enabling academic and industrial communities to analyze trade-offs, identify bottlenecks, and optimize for energy efficiency.

The landscape of LLM inference is rapidly evolving, and TokenPowerBench is designed to evolve with it. We will extend coverage beyond our current H100 testbed to other GPU architectures, including the next NVIDIA generations and AMD accelerators, as well as emerging AI chips and DPUs. We will also polish the benchmark and extend the functionality to quantify the trade-off between inference accuracy and energy efficiency, providing users with guidance on where energy savings begin to erode model quality.

% We invite the broader community to adopt, extend, and contribute to TokenPowerBench at \url{https://github.com/TokenPowerBench}.

% \include{1_introduction}
% \include{2_background}
% \include{3_newmethod}
% \include{4_evaluation}
% \include{5_relatedwork}
% \include{6_conclusion}

% \nocite{*}
\section{Acknowledgments}

We are thankful to the anonymous reviewers for their valuable feedback. This research is supported in part by the National Science Foundation under grant OAC-2404438 and CNS-1939140 (A U.S. National Science Foundation Industry-University Cooperative Research Center on Cloud and Autonomic Computing). We are also very grateful to the High Performance Computing Center of Texas Tech University for providing HPC resources for this project.

\bigskip
\bibliography{aaai2026}

\end{document}